\documentclass[runningheads]{llncs}
\usepackage{graphicx}

\usepackage{tikz}
\usepackage{comment} 
\usepackage{amsmath,amssymb} 
\usepackage{color}
\usepackage[breaklinks=true]{hyperref}
\graphicspath{ {images/} }

\begin{document}
\pagestyle{headings}
\mainmatter
\def\ECCVSubNumber{2992}  

\title{Reconstructing NBA Players} 

\titlerunning{Reconstructing NBA Players}
%
\author{Luyang Zhu \and
Konstantinos Rematas \and
Brian Curless \and \\
Steven M. Seitz \and
Ira Kemelmacher-Shlizerman}

%
\authorrunning{Zhu et al.}
%
\institute{University of Washington}

\maketitle


\begin{abstract}

Great progress has been made in 3D body pose and shape estimation from a single photo.  Yet, state-of-the-art results still suffer from errors due to challenging body poses, modeling clothing, and self occlusions. The domain of basketball games is particularly challenging, as it exhibits all of these challenges. In this paper, we introduce a new approach for reconstruction of basketball players that outperforms the state-of-the-art.  Key to our approach is a new method for creating poseable, skinned models of NBA players, and a large database of meshes (derived from the NBA2K19 video game) that we are releasing to the research community. Based on these models, we introduce a new method that takes as input a single photo of a clothed player in any basketball pose and outputs a high resolution mesh and 3D pose for that player. We demonstrate substantial improvement over state-of-the-art, single-image methods for body shape reconstruction. Code and dataset are available at \url{http://grail.cs.washington.edu/projects/nba\_players/}.


\end{abstract}

\section{Introduction}

Given regular, broadcast video of an NBA basketball game, we seek a complete 3D reconstruction of the players, viewable from any camera viewpoint.  This reconstruction problem is challenging for many reasons, including the need to infer hidden and back-facing surfaces, and the complexity of basketball poses, e.g., reconstructing jumps, dunks, and dribbles. 

Human body modeling from images has advanced dramatically in recent years, due in large part to availability of 3D human scan datasets, e.g., CAESAR \cite{robinette2002civilian}.  Based on this data, researchers have developed powerful tools that enable recreating realistic humans in a wide variety of poses and body shapes \cite{loper2015smpl}, and estimating 3D body shape from single images \cite{pifuSHNMKL19,weng2019photo}.  These models, however, are largely limited to the domains of the source data -- people in underwear \cite{robinette2002civilian}, or clothed models of people in static, staged poses \cite{renderpeople}. Adapting this data to a domain such as basketball is extremely challenging, as we must not only match the physique of an NBA player, but also their unique basketball poses.

\begin{figure}
    \centering
    \includegraphics[width=1.0\textwidth]{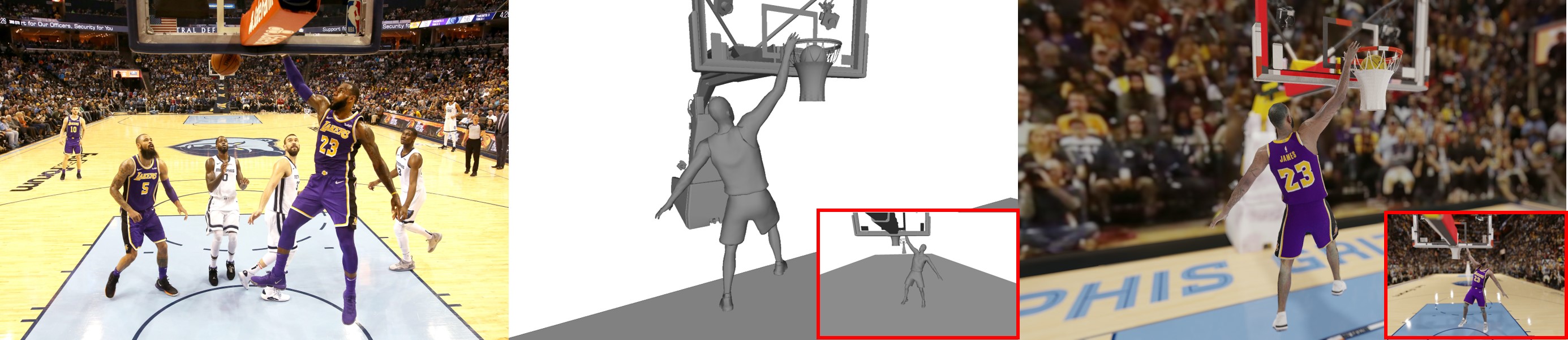}
    \caption{Single input photo (left), estimated 3D posed model that is viewed from \textbf{a new} camera position (middle), same model with video game texture for visualization purposes. The insets show the estimated shape from the input camera viewpoint. (Court and basketball meshes are extracted from the video game) {\em Photo Credit:~\cite{usatoday}}}
    \label{fig:teaser}
\end{figure}

Sports video games, on the other hand, have become extremely realistic, with renderings that are increasingly difficult to distinguish from reality. The player models in games like NBA2K \cite{nba2k} are meticulously crafted to capture each player's physique and appearance (Fig.~\ref{training_data}).  Such models are ideally suited as a training set for 3D reconstruction and visualization of real basketball games.

In this paper, we present a novel dataset and  neural networks that reconstruct high quality meshes of basketball players and retarget these meshes to fit frames of real NBA games. Given an image of a player, we are able to reconstruct the action in 3D, and apply new camera effects such as close-ups, replays, and bullet-time effects (Fig.~\ref{fig:teaser}).

Our new dataset is derived from the video game NBA2K (with approval from the creator, Visual Concepts), by playing the game for hours and intercepting rendering instructions to capture thousands of meshes in diverse poses. Each mesh provides detailed shape and texture, down to the level of wrinkles in clothing, and captures all sides of the player, not just those visible to the camera. Since the intercepted meshes are not rigged, we learn a mapping from pose parameters to mesh geometry with a novel {\em deep skinning} approach.  The result of our skinning method is a detailed deep net basketball body model that can be retargeted to any desired player and basketball pose.

We also introduce a system to fit our retargetable player models to real NBA game footage by solving for 3D player pose and camera parameters for each frame.  We demonstrate the effectiveness of this approach on synthetic and real NBA input images, and compare with the state of the art in 3D pose and human body model fitting. Our method outperforms the state-of-the-art methods when reconstructing basketball poses and players even when these methods, to the extent possible, are retrained on our new dataset. This paper focuses on basketball shape estimation, and leaves texture estimation as future work. 

Our biggest contributions are, first, a deep skinning approach that produces high quality, pose-dependent models of NBA players. A key differentiator is that we leverage {\em thousands of poses} and capture detailed geometric variations as a function of pose (e.g., folds in clothing), rather than a small number of poses which is the norm for datasets like CAESAR (1-3 poses/person) and modeling methods like SMPL (trained on CAESAR and $\sim$45 poses/person). 
While our approach is applicable to any source of registered 3D scan data, 
we apply it to reconstruct models of NBA players from NBA2K19 game play screen captures.  As such, a second key contribution is pose-dependent models of different basketball players, and raw capture data for the research community. Finally, we present a system that fits these player models to images, enabling 3D reconstructions from photos of NBA players in real games.
Both our skinning and pose networks are evaluated quantitatively and qualitatively, and outperform the current state of the art.

One might ask, why spend so much effort reconstructing mesh models that already exist (within the game)? NBA2K's rigged models and in-house animation tools are proprietary IP.  By reconstructing a posable model from intercepted meshes (eliminating requirement of proprietary animation and simulation tools), we can provide these best-in-the-world models of basketball players to researchers for the first time (with the company's support).  These models provide a number of advantages beyond existing body models such as SMPL.  In particular, they capture not just static poses, but human body dynamics for running, walking, and many other challenging activities.  Furthermore, the plentiful pose-dependent data enables robust reconstruction even in the presence of heavy occlusions.  In addition to producing the first high quality reconstructions of basketball from regular photos, our models can facilitate synthetic data collection for ML algorithms.  Just as simulation provides a critical source of data for many ML tasks in robotics, self-driving cars, depth estimation, etc., our derived models can generate much more simulated content under any desired conditions (we can render any pose, viewpoint, combination of players, against any background, etc.) 


\begin{figure*}
\begin{center}
   \includegraphics[width=1.0\linewidth]{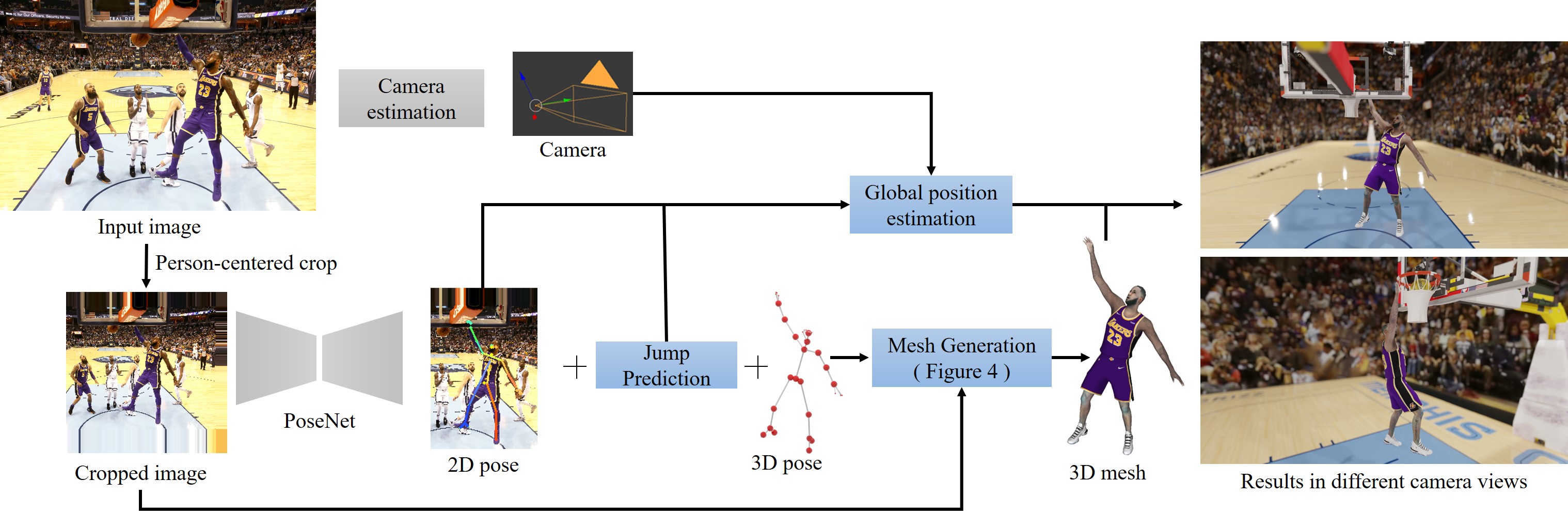}
\end{center}
   \caption{Overview: Given a single basketball image (top left), we begin by  detecting the target player using \cite{cao2018openpose,simon2017hand}, and create a person-centered crop (bottom left).  From this crop, our PoseNet predicts 2D pose, 3D pose, and jump information. The estimated 3D pose and the cropped image are then passed to mesh generation networks to predict the full, clothed 3D mesh of the target player. Finally, to globally position the player on the 3D court (right), we estimate camera parameters by solving the PnP problem on known court lines and predict global player  position by combining camera, 2D pose, and jump information. Blue boxes represent novel components of our method.
} 
\label{pipeline}

\end{figure*}

\section{Related Work}
\noindent{\bf{Video Game Training Data.}}
Recent works \cite{Richter_2016_ECCV,Richter_2017,Krahenbuhl18,rematas2018soccer} 
have shown that, for some domains, data derived from video games can significantly reduce manual labor and labeling, since ground-truth labels can be extracted automatically while playing the game. E.g., \cite{calagari15,rematas2018soccer} collected depth maps of soccer players by playing the FIFA soccer video game, showing generalization to images of real games. Those works, however, focused on low level vision data, e.g., optical flow and depth maps rather than full high quality meshes. In contrast, we collect data that includes 3D triangle meshes, texture maps, and detailed 3D body pose, which requires more sophisticated modeling of human body pose and shape.

\noindent{\bf{Sports 3D reconstruction.}}
Reconstructing 3D models of athletes playing various sports from images has been explored in both academic research and industrial products. Most previous methods use multiple camera inputs rather than a single view. Grau \textit{et al.} \cite{grau2007robust,grau2007free} and Guillemaut \textit{et al.} \cite{Guillemaut09,Guillemaut2011} used multiview stereo methods for free viewpoint navigation. Germann \textit{et al.} \cite{germann2010articulated} proposed an articulated billboard presentation for novel view interpolation. Intel demonstrated $360$ degree viewing experiences\footnote{\url{https://www.intel.com/content/www/us/en/sports/technology/true-view.html}}, with their True View~\cite{inteltrueview} technology by installing $38$ synchronized 5k cameras around the venue and using this multi-view input to build a volumetric reconstruction of each player. This paper aims to achieve similar reconstruction quality but from a \textit{single} image. 

Rematas \textit{et al.}~\cite{rematas2018soccer}  reconstructed soccer games from monocular YouTube videos. However, they predicted only depth maps, thus can not handle occluded body parts
and player visualization from all angles.
Additionally, they estimated players' global position by assuming all players are standing on the ground, which is not a suitable assumption for basketball, where players are often airborne. The detail of the depth maps is also low. We address all of these challenges by building a basketball specific player reconstruction algorithm that is trained on meshes and accounts for complex airborne basketball poses. Our result is a detailed mesh of the player from a single view, but comparable to multi-view reconstructions. Our reconstructed mesh can be viewed from any camera position. 

\noindent{\bf{3D human pose estimation.}}
Large scale body pose estimation datasets\\ \cite{ionescu2013human3,mehta2017monocular,von2018recovering} enabled great progress in 3D human pose estimation from single images\\ \cite{mehta2017vnect,martinez2017simple,sun2018integral,habibieCVPR19,Moon_2019_ICCV_3DMPPE}. 
We build on \cite{mehta2017vnect} but train on our new basketball pose data, use a more detailed skeleton (35 joints including fingers and face keypoints), and an explicit model of jumping and camera to predict global position. Accounting for jumping is an important step that allows our method outperform state of the art pose. 

\noindent{\bf{3D human body shape reconstruction.}}
Parametric human body models \cite{anguelov2005scape,loper2015smpl,Dyna:SIGGRAPH:2015,MANO:SIGGRAPHASIA:2017,joo2018total,pavlakos2019smlx} are commonly fit to images to derive a body skeleton, and provide a framework to optimize for shape parameters \cite{Bogo:ECCV:2016,joo2018total,pavlakos2019smlx,xiang2019monocular,lassner2017unite,huang2017towards,zanfir2018monocular}. \cite{weng2019photo} further 2D warped the optimized parametric model to approximately account for clothing and create a rigged animated mesh from a single photo.  \cite{hmrKanazawa17,pavlakos2018learning,humanMotionKanazawa19,kolotouros2019cmr,pavlakos2019texturepose,Guler_2019_CVPR,Zhu_2019_CVPR,kolotouros2019spin} trained a neural network to directly regress body shape parameters from images. Most parametric model based methods reconstruct undressed humans, since clothing is not part of the parametric model.

Clothing can be modeled to some extent by warping SMPL~\cite{loper2015smpl} models, e.g., to silhouettes: Weng \textit{et al.} \cite{weng2019photo} demonstrated 2D warping of depth and normal maps from a single photo silhouette, and Alldeick \textit{et al.} \cite{alldieck2018video,alldieck19cvpr,alldieck2019tex2shape} addressed multi-image fitting. Alternatively, given predefined garment models \cite{bhatnagar2019mgn} estimated a clothing mesh layer on top of SMPL. 

Non-parametric methods~\cite{varol18_bodynet,natsume2019siclope,pifuSHNMKL19,pumarola20193dpeople} proposed voxel~\cite{varol18_bodynet} or implicit function~\cite{pifuSHNMKL19} representations to model  clothed humans by training on representative synthetic data. Xu \textit{et al.} \cite{Xu-Video-based-Characters,XuMonoPerfCap} and Habermann \textit{et al.} \cite{habermann2019TOG} assumed a pre-captured multi-view model of the clothed human, retargeted based on new poses. 

We focus on single-view reconstruction of players in NBA basketball games, producing a complete 3D model of the player pose and shape, viewable from any camera viewpoint.  This reconstruction problem is challenging for many reasons, including the need to infer hidden and back-facing surfaces, and the complexity of basketball poses, e.g., reconstructing jumps, dunks, and dribbles. 
Unlike prior methods modeling undressed people in various poses or dressed people in a frontal pose, we focus on modeling clothed people under challenging basketball poses and provide a rigorous comparison with the state of the art.

\section{The NBA2K Dataset} \label{sec:dataset}

\begin{figure}
\begin{center}
	\includegraphics[width=1.0\linewidth]{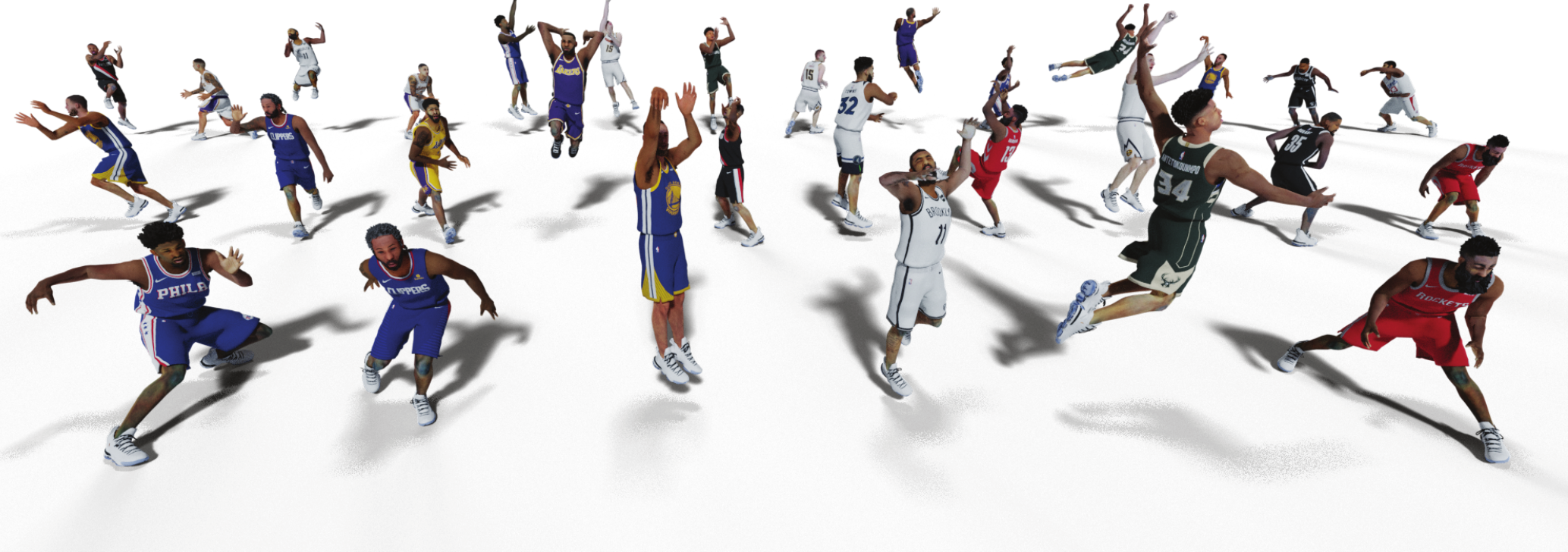}
\end{center}
   \caption{Our novel NBA2K dataset examples, extracted from the NBA2K19 video game. Our NBA2K dataset captures 27,144 basketball poses spanning 27 subjects, extracted from the NBA2K19 video game.}
\label{training_data}
\end{figure}

Imagine having thousands of 3D body scans of NBA players, in every conceivable pose during a basketball game.
Suppose that these models were extremely detailed and realistic, down to the level of wrinkles in clothing.
Such a dataset would be instrumental for sports reconstruction, visualization, and analysis.
This section describes such a dataset, which we call \textit{NBA2K}, after the video game from which these models derive. 
These models of course are not literally player scans, but are produced by professional modelers for use in the NBA2K19 video game, based on a variety of data including high resolution player photos, scanned models and mocap data of some players. While they do not exactly match each player, they are among the most accurate 3D renditions in existence
(Fig.~\ref{training_data}).

Our NBA2K dataset consists of body mesh and texture data for several NBA players, each in around 1000 widely varying poses. For each mesh (vertices, faces and texture) we also provide its 3D pose (35 keypoints including face and hand fingers points) and the corresponding RGB image with its camera parameters. While we used meshes of 27 real famous players to create many of figures in this paper, we do not have permission to release models of current NBA players. Instead, we additionally collected the same kind of data for 28 synthetic players and retrained our pipeline on this data. The synthetic player's have the same geometric and visual quality as the NBA models and their data along with trained models will be shared with the research community upon publication of this paper. Our released meshes, textures, and models will have the same quality as what's in the paper, and span a similar variety of player types, but not be named individuals. Visual Concepts~\cite{nba2k} has approved our collection and sharing of the data.

The data was collected by playing the NBA2K19 game and intercepting calls between the game engine and the graphics card using RenderDoc~\cite{Renderdoc}.  The program captures all drawing events per frame, where we locate player rendering events by analyzing the hashing code of both vertex and pixel shaders. Next, triangle meshes and textures are extracted by reverse-engineering the compiled code of the vertex shader. The game engine renders players by body parts, so we perform a nearest neighbor clustering to decide which body part belongs to which player. Since the game engine optimizes the mesh for real-time rendering, the extracted meshes have different mesh topologies, making them harder to use in a learning framework. We register the meshes by resampling vertices in texture space based on a template mesh. After registration, the processed mesh has 6036 vertices and 11576 faces with fixed topology across poses and players (point-to-point correspondence), has multiple connected components (not a watertight manifold), and comes with no skinning information. We also extract the rest-pose skeleton and per-bone transformation matrix, from which we can compute forward kinematics to get full 3D pose. 

\section{From Single Images to Meshes}
\label{sec:method}

Figure~\ref{pipeline} shows our full reconstruction system, starting from a single image of a basketball game, and ending with output of a complete, high quality mesh of the target player with pose and shape matching the image. Next, we describe the individual steps to achieve the final results.

\subsection{3D Pose in World Coordinates}

\noindent{\bf{2D pose, jump, and 3D pose estimation}} Since our input meshes are not rigged (no skeletal information or blending weights), we propose a neural network called {\em PoseNet} to estimate the 3D pose and other attributes of a player from a single image.  This 3D pose information will be used later to facilitate shape reconstruction. PoseNet takes a single image as input and is trained to output 2D body pose, 3D body pose, a binary jump classification (is the person airborne or not), and the jump height (vertical height of the feet from ground).  The two jump-related outputs are key for global position estimation and are our novel addition to existing generic body pose estimation.

From the input image, we first extract ResNet~\cite{xiao2018simple} features (from layer 4) and supply them to four separate network branches. The output of the 2D pose branch is a set of 2D heatmaps (one for each 2D keypoint) indicating where the particular keypoint is located. The output of the 3D pose branch is a set of $XYZ$ location maps (one for each keypoint)~\cite{mehta2017vnect}. The location map indicates the possible 3D location for every pixel. The 2D and 3D pose branches use the same architecture as~\cite{xiao2018simple}.
The \textit{jump branch} estimates a class label, and the \textit{jump height branch} regresses the height of the jump. Both networks use a fully connected layer followed by two linear residual blocks~\cite{martinez2017simple} to get the final output.

The PoseNet model is trained using the following loss:
\begin{equation}
\begin{split}
    \mathcal{L}_{pose} &= \omega_{2d}\mathcal{L}_{2d} + \omega_{3d}\mathcal{L}_{3d} + \omega_{bl}\mathcal{L}_{bl}+\omega_{jht}\mathcal{L}_{jht} + \omega_{jcls}\mathcal{L}_{jcls}
\end{split}
\end{equation}
where 
$\mathcal{L}_{2d}=\lVert {H - \hat{H}} \rVert_{1}$ is the loss between predicted ($H$) and ground truth ($\hat{H}$) heatmaps,
$\mathcal{L}_{3d}=\lVert {L - \hat{L}} \rVert_{1}$ is the loss between predicted ($L$) and ground truth ($\hat{L}$) 3D location maps,
$\mathcal{L}_{bl}=\lVert {B - \hat{B}} \rVert_{1}$ is the loss between predicted ($B$) and ground truth ($\hat{B}$) bone lengths to penalize unnatural 3D poses (we pre-computed the ground truth bone length over the training data),
$\mathcal{L}_{jht} =\lVert {h - \hat{h}} \rVert_{1}$ is the loss between predicted ($h$) and ground truth ($\hat{h}$) jump height, and
$\mathcal{L}_{jcls}$ is the cross-entropy loss for the jump class.
For all experiments, we set $\omega_{2d}=10,~\omega_{3d}=10,~\omega_{bl}=0.5,~\omega_{jht}=0.4$, and $\omega_{jcls}=0.2$.

\noindent{\bf{Global Position}} To estimate the global position of the player we need the camera parameters of the input image. 
Since NBA courts have known dimensions, we generate a synthetic 3D field and align it with the input frame. Similar to \cite{rematas2018soccer,CarrPointless2012}, we use a two-step approach. First, we provide four manual correspondences between the input image and the 3D basketball court to initialize the camera parameters by solving PnP~\cite{lepetit2009epnp}.
Then, we perform a line-based camera optimization similar to~\cite{rematas2018soccer}, where the projected lines from the synthetic 3D court should match the lines on the image. Given the camera parameters, we can estimate a player's global position on (or above) the 3D court by the lowest keypoint and the jump height. We cast a ray from the camera center through the image keypoint; the 3D location of that keypoint is where the ray-ground height is equal to the estimated jump height.

\subsection{Mesh Generation}
\label{sec:skinning}
\begin{figure}
\begin{center}
	\includegraphics[width=1.0\linewidth]{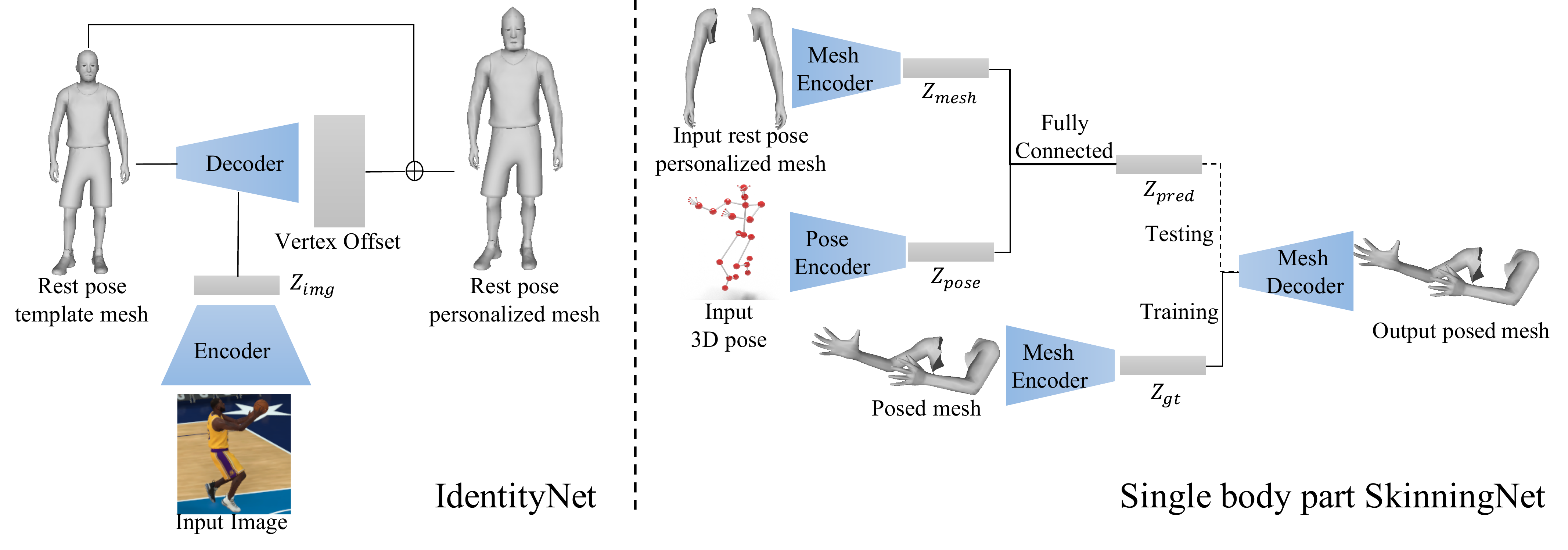}
\end{center}

   \caption{Mesh generation contains two sub networks: IdentityNet and SkinningNet.  IdentityNet deforms a rest pose template mesh (average rest pose over all players in the database), into a rest pose personalized mesh given the image. SkinningNet takes the rest pose personalized mesh and 3D pose as input and outputs the posed mesh. There is a separate SkinningNet per body part, here we illustrate the arms.}
\label{meshGeneration}
\end{figure}

Reconstruction of a complete detailed 3D mesh (including deformation due to pose, cloth, fingers and face) from a single image is a key technical contribution of our method. To achieve this we introduce two sub-networks (Fig.~\ref{meshGeneration}):  {\em IdentityNet} and {\em SkinningNet}.  IdentityNet takes as input an image of a player whose rest mesh we wish to infer, and outputs the person's rest mesh by deforming a template mesh. The template mesh is the average of all training meshes and is the same starting point for any input. The main benefit of this network is that it allows us to estimate the body size and arm span of the player according to the input image. SkinningNet takes the rest pose personalized mesh and the 3D pose as input, and outputs the posed mesh. To reduce the learning complexity, we pre-segment the mesh into six parts: head, arms, shirt, pants, legs and shoes. We then train a SkinningNet on each part separately. Finally, we combine the six reconstructed parts into one, while removing interpenetration of garments with body parts. Details are described below. 

\noindent{\bf{IdentityNet.}} 
We propose a variant of 3D-CODED~\cite{groueix20183d} to deform the template mesh. We first use ResNet~\cite{he2016deep} to extract features from input images. Then we concatenate template mesh vertices with image features and send them into an AtlasNet decoder~\cite{groueix2018papier} to predict per vertex offsets. Finally, we add this offset to the template mesh to get the predicted personalized mesh. We use the L1 loss between the prediction and ground truth to train IdentityNet.

\noindent{\bf{SkinningNet.}} 
We propose a TL-embedding network~\cite{girdhar2016learning} to learn an embedding space with generative capability.
Specifically, the 3D keypoints $K_{pose} \in R^{35\times3}$ are processed by the pose encoder to produce a latent code $Z_{pose} \in R^{32}$. The rest pose personalized mesh vertices $V_{rest} \in R^{N\times3}$(where $N$ is the number of vertices in a mesh part) are processed by the mesh encoder to produce a latent code $Z_{rest}\in R^{32}$. Then $Z_{pose}$ and $Z_{rest}$ are concatenated and fed into a fully connected layer to get $Z_{pred}\in R^{32}$. Similarly, the ground truth posed mesh vertices $V_{posed} \in R^{N \times 3}$ are processed by another mesh encoder to produce a latent code $Z_{gt}\in R^{32}$. $Z_{gt}$ is sent into the mesh decoder during training while $Z_{pred}$ is sent into the mesh decoder during testing. 

The Pose encoder is comprised of two linear residual blocks~\cite{martinez2017simple} followed by a fully connected layer. The mesh encoders and shared decoder are built with spiral convolutions~\cite{bouritsas2019neural}. See supplementary material for detailed network architecture. 
SkinningNet is trained with the following loss:
\begin{equation}
\mathcal{L}_{skin} = \omega_{Z}\mathcal{L}_{Z}+\omega_{mesh}\mathcal{L}_{mesh}
\end{equation}
where $\mathcal{L}_{Z}=\lVert {Z_{pred} - Z_{gt}} \rVert_{1}$ forces the space of $Z_{pred}$ and $Z_{gt}$ to be similar, and $\mathcal{L}_{mesh}=\lVert {V_{pred} - V_{posed}}\rVert_{1}$ is the loss between decoded mesh vertices $V_{pred}$ and ground truth vertices $V_{posed}$. The weights of different losses are set to $\omega_{Z}=5,~\omega_{mesh}=50$. See supplementary for detailed training parameters.

\noindent{\bf{Combining body part meshes.}} Direct concatenation of body parts results in interpenetration between the garment and the body. Thus, we first detect all body part vertices in collision with clothing as in~\cite{pavlakos2019smlx}, and then follow \cite{sorkine2007rigid,sorkine2004laplacian} to deform the mesh by moving collision vertices inside the garment while preserving local rigidity of the mesh. This detection-deformation process is repeated until there is no collision or the number of iterations is above a threshold (10 in our experiments). See supplementary material for details of the optimization. 


\begin{table}
\begin{center}
\begin{tabular}{c|c|c|c|c|c|c}
\hline
& HMR~\cite{hmrKanazawa17} & CMR~\cite{kolotouros2019cmr} & SPIN~\cite{kolotouros2019spin} & Ours(Reg+BL) & Ours(Loc) & Ours(Loc+BL) \\
\hline
MPJPE & 115.77 & 82.28 & 88.72 & 81.66 & 66.12 & \textbf{51.67}\\
\hline
MPJPE-PA & 78.17 & 61.22 & 59.85 & 63.70 & 52.73 & \textbf{40.91}\\
\hline
\end{tabular}
\end{center}
\caption{\textbf{Quantitative comparison of 3D pose estimation to state of the art.} The metric is mean per joint position error with (MPJPE-PA) and without (MPJPE) Procrustes alignment. Baseline methods are fine-tuned on our NBA2K dataset. }
\label{pose_table}
\end{table}

\begin{table}
\begin{center}
\begin{tabular}{c|c|c|c|c|c}
\hline
& HMR~\cite{hmrKanazawa17}  & SPIN~\cite{kolotouros2019spin} & SMPLify-X~\cite{pavlakos2019smlx} & PIFu~\cite{pifuSHNMKL19} & Ours \\
\hline
CD &  22.411 & 14.793 & 47.720 & 23.136 & \textbf{4.934}\\
\hline
EMD &  0.137 & 0.125 & 0.187 & 0.207 & \textbf{0.087}\\
\hline
\end{tabular}
\end{center}
\caption{\textbf{Quantitative comparison of our mesh reconstruction to state of the art.} We use Chamfer distance denoted by CD (scaled by 1000, lower is better), and Earth-mover distance denoted by EMD (lower is better) for comparison. Both distance metrics show that our method significantly outperforms state of the art for mesh estimation. All related works are retrained or fine-tuned on our data, see text.}
\label{mesh_table}
\end{table}

\section{Experiments}
\noindent{\bf{Dataset Preparation.}} We evaluate our method with respect to the state of the art on our NBA2K dataset. We collected 27,144 meshes spanning 27 subjects performing various basketball poses (about 1000 poses per player). PoseNet training requires generalization on real images. Thus, we augment the data to 265,765 training examples, 37,966 validation examples, and  66,442 testing examples. Augmentation is done by rendering and blending meshes into various random basketball courts. For IdentityNet and SkinningNet,  we select 19,667 examples from 20 subjects as training data and test on 7,477 examples from 7 unseen players. To further evaluate generalization of our method, we also provide qualitative results on real images. Note that textures are extracted from the game and not estimated by our algorithm. 

\subsection{3D Pose, Jump, and Global Position Evaluation}
We evaluate pose estimation by comparing to state of the art SMPL-based methods that released training code. Specifically we compare with HMR~\cite{hmrKanazawa17}, CMR~\cite{kolotouros2019cmr}, and SPIN~\cite{kolotouros2019spin}. For fair comparison, we fine-tuned their models with 3D and 2D ground-truth NBA2K poses. Since NBA2K and SMPL meshes have different topology we do not use mesh vertices and SMPL parameters as part of the supervision.  Table~\ref{pose_table} shows comparison results for 3D pose. The metric is defined as mean per joint position error (MPJPE) with and without procrustes alignment. The error is computed on 14 joints as defined by the LSP dataset~\cite{Johnson10}.  Our method outperforms all other methods even when they are fine-tuned on our NBA2K dataset (lower number is better).  

To further evaluate our design choices, we compare the location-map-based representation (used in our network) with direct regression of 3D joints, and also evaluate the effect of bone length (BL) loss on pose prediction. A direct regression baseline is created by replacing our deconvolution network with fully connected layers~\cite{martinez2017simple}. The effectiveness of BL loss is evaluated by running the network with and without it. As shown in Table~\ref{pose_table}, both location maps and BL loss can boost the performance.
In supplementary material, we show our results on global position estimation. We can see that our method can accurately place players (both airborne and on ground) on the court due to accurate jump estimation.

\begin{figure}
\begin{center}
  \includegraphics[width=1.0\linewidth]{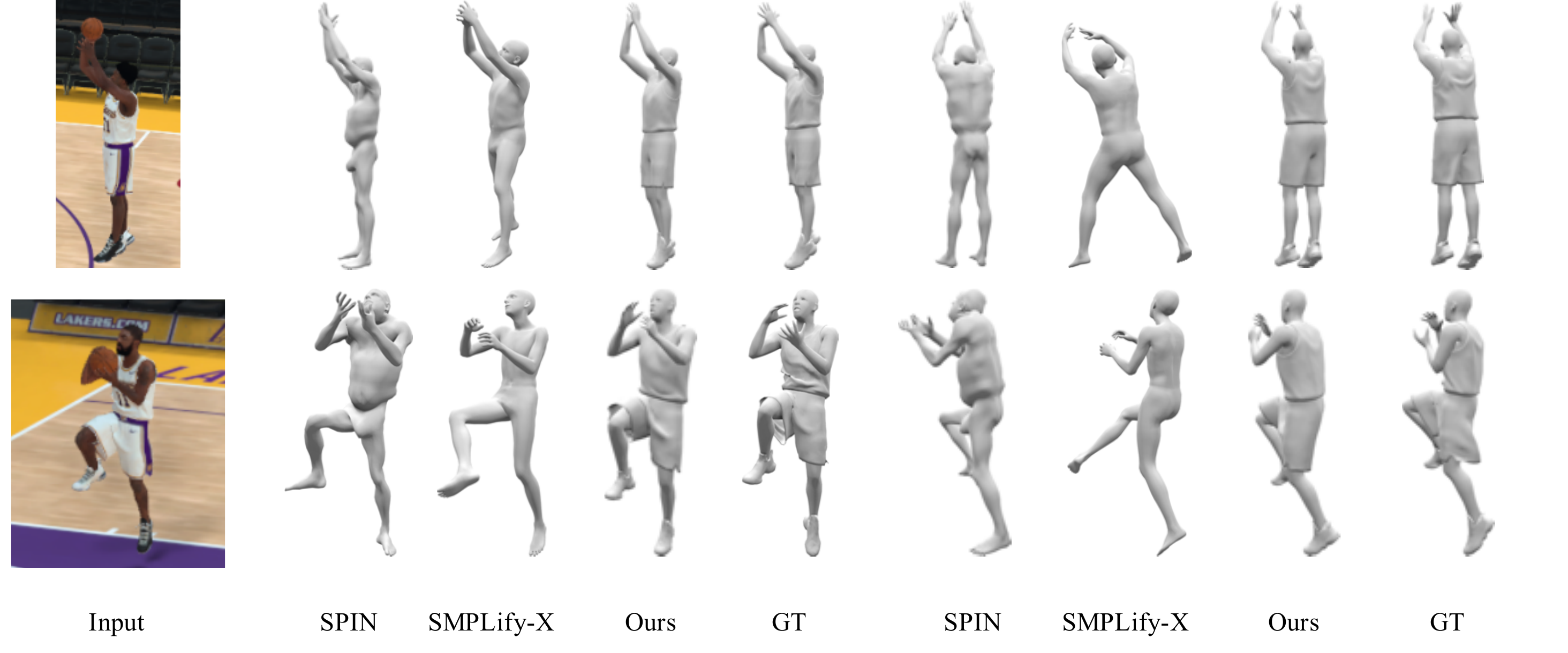}
\end{center}
  \caption{\textbf{Comparison with SMPL-based methods.} Column 1 is input, columns 2-5 are reconstructions in the image view, columns 6-9 are visualizations from a novel view. Note the significant difference in body pose between ours and SMPL-based methods.}
\label{smpl_shape}
\end{figure}

\begin{figure}
\begin{center}
   \includegraphics[width=1\linewidth]{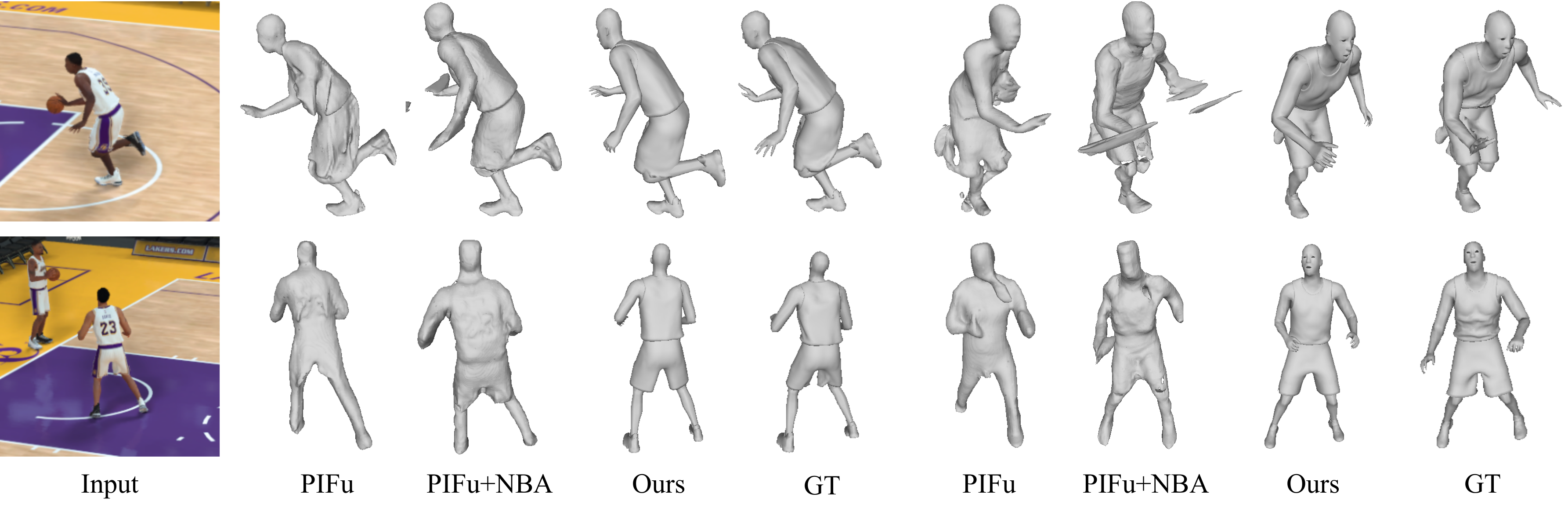}
\end{center}
   \caption{\textbf{Comparison with PIFu\cite{pifuSHNMKL19}.} Column 1 is input, columns 2-5 are reconstructions in the image viewpoint, columns 6-9 are visualizations from a novel view. PIFu significantly over-smooths shape details and produces lower quality reconstruction even when trained on our dataset (PIFu+NBA).} 
\label{pifu}
\end{figure}

\begin{figure}
\begin{center}
  \includegraphics[width=1\linewidth]{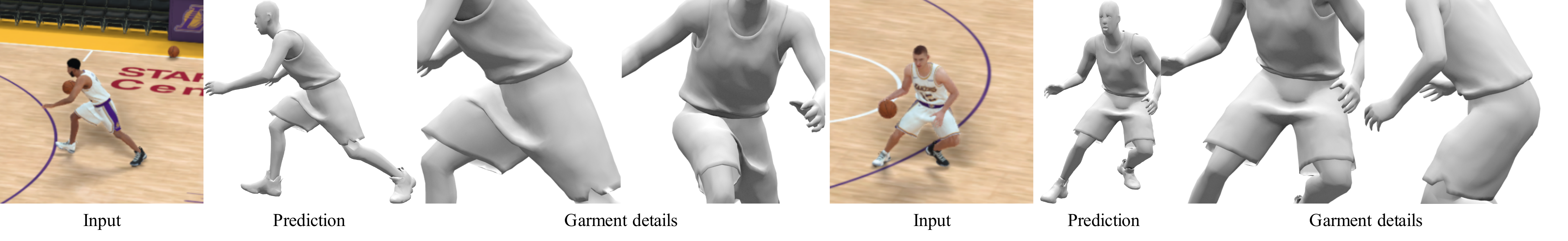}
\end{center}

  \caption{\textbf{Garment details at various poses.} For each input image, we show the predicted shape, close-ups from two viewpoints.} 
\label{shirt_wrinkles}
\end{figure}

\subsection{3D Mesh Evaluation}
\noindent{\bf{Quantitative Results.}} Table~\ref{mesh_table} shows results of comparing our mesh reconstruction method to the state of the art on NBA2K data. We compare to both undressed (HMR~\cite{hmrKanazawa17}, SMPLify-X~\cite{pavlakos2019smlx}, SPIN~\cite{kolotouros2019spin}) and clothed (PIFu~\cite{pifuSHNMKL19}) human reconstruction methods. For fair comparison, we retrain PIFU on our NBA2K meshes. SPIN and HMR are based on the SMPL model where we do not have groundtruth meshes, so we fine-tuned with NBA2K 2D and 3D pose. SMPLify-X is an optimization method, so we directly apply it to our testing examples. The meshes generated by baseline methods and the NBA2K meshes do not have one-to-one vertex correspondence, thus we use Chamfer (CD) and Earth-mover (EMD) as distance metrics. 
Prior to distance computations, all predictions are aligned to ground-truth using ICP. We can see that our method outperforms both undressed and clothed human reconstruction methods even when they are trained on our data.

\begin{figure}
\begin{center}
  \includegraphics[width=1\linewidth]{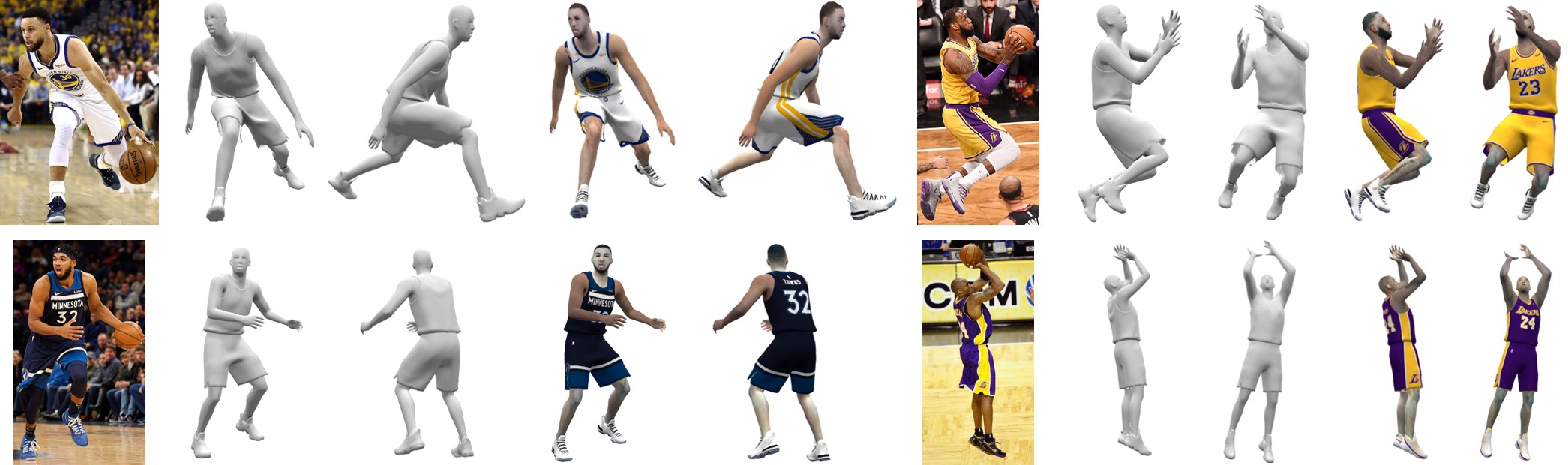}
\end{center}
  \caption{\textbf{Results on real images.} For each example, column 1 is the input image, 2-3 are reconstructions rendered in different views. 4-5 are corresponding renderings using texture from the video game, just for visualization. Our technical method is focused only on shape recovery. {\em Photo Credit:~\cite{getty}}} 
\label{real_image}
\end{figure}

\noindent{\bf{Qualitative Results.}} Fig.~\ref{smpl_shape} qualitatively compares our results with the best performing SMPL-based methods SPIN~\cite{kolotouros2019spin} and SMPLify-X~\cite{pavlakos2019smlx}. These two methods do not reconstruct clothes, so we focus on the pose accuracy of the body shape. Our method generates more accurate body shape for basketball poses, especially for hands and fingers. Fig.~\ref{pifu} qualitatively compares with  PIFu~\cite{pifuSHNMKL19}, a state-of-the-art {\em clothed} human reconstruction method.  Our method  generates detailed geometry such as shirt wrinkles under different poses while PIFu tends to over-smooth faces, hands, and garments. Fig.~\ref{shirt_wrinkles} further visualizes garment details in our reconstructions. Fig.~\ref{real_image} shows results of our  method on real images, demonstrating robust generalization. Please also refer to the supplementary pdf and video for high quality reconstruction of real NBA players.

\subsection{Ablative Study}

\begin{figure}
\begin{center}
  \includegraphics[width=0.8\linewidth]{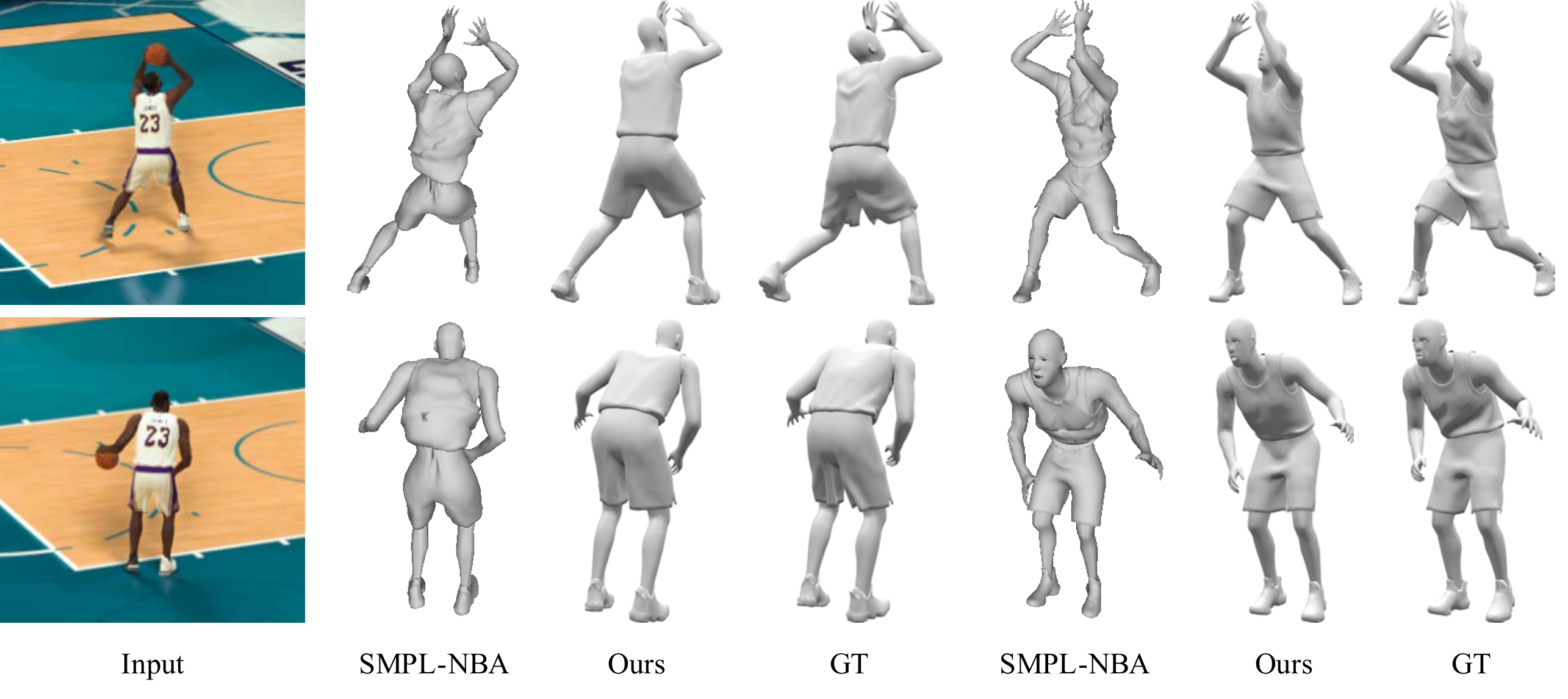}
\end{center}

  \caption{\textbf{Comparison with SMPL-NBA.} 
  Column 1 is input, columns 2-4 are reconstructions in the image view, columns 5-7 are visualizations from a novel viewpoint. SMPL-NBA fails to model clothing and the fitting process is unstable. } 
\label{smpl_nba}
\end{figure}

\noindent{\bf{Comparison with SMPL-NBA.}} We follow the idea of SMPL~\cite{loper2015smpl} to train a skinning model from NBA2K registered mesh sequences. The trained body model is called SMPL-NBA. Since we don't have rest pose meshes for thousands of different subjects, we cannot learn a meaningful PCA shape basis as SMPL did. Thus, we focus on the pose dependent part and fit the SMPL-NBA model to 2000 meshes of a single player. We use the same skeleton rig as SMPL to drive the mesh. Since our mesh is comprised of  multiple connected parts, we initialize the skinning weights using a voxel-based heat diffusion method~\cite{dionne2013geodesic}. The whole training process of SMPL-NBA is the same as the pose parameter training of SMPL. We fit the learned model to predicted 2D keypoints and 3D keypoints from PoseNet following SMPLify~\cite{Bogo:ECCV:2016}. Fig.~\ref{smpl_nba} compares SkinningNet with SMPL-NBA, showing that SMPL-NBA has severe artifacts for garment deformation -- an inherent difficulty for traditional skinning methods. It also suffers from twisted joints which is a common problem when fitting per bone transformation to 3D and 2D keypoints. 

\begin{table}
\begin{center}
\begin{tabular}{c|c|c|c}
\hline
& CMR~\cite{kolotouros2019cmr}  & 3D-CODED~\cite{groueix20183d} & Ours \\
\hline
MPVPE &  85.26 & 84.22 & \textbf{76.41}\\
\hline
MPVPE-PA &  64.32 & 63.13 & \textbf{54.71}\\
\hline
\end{tabular}
\end{center}
\caption{\textbf{Quantitative comparison with 3D-CODED~\cite{groueix20183d} and CMR~\cite{kolotouros2019cmr}.} The metric is mean per vertex position error in mm with (MPVPE-PA) and without (MPVPE) Procrustes alignment. All baseline methods are trained on the NBA2K data.}
\label{geometry_table}
\end{table}

\begin{figure}
\begin{center}
  \includegraphics[width=1.0\linewidth]{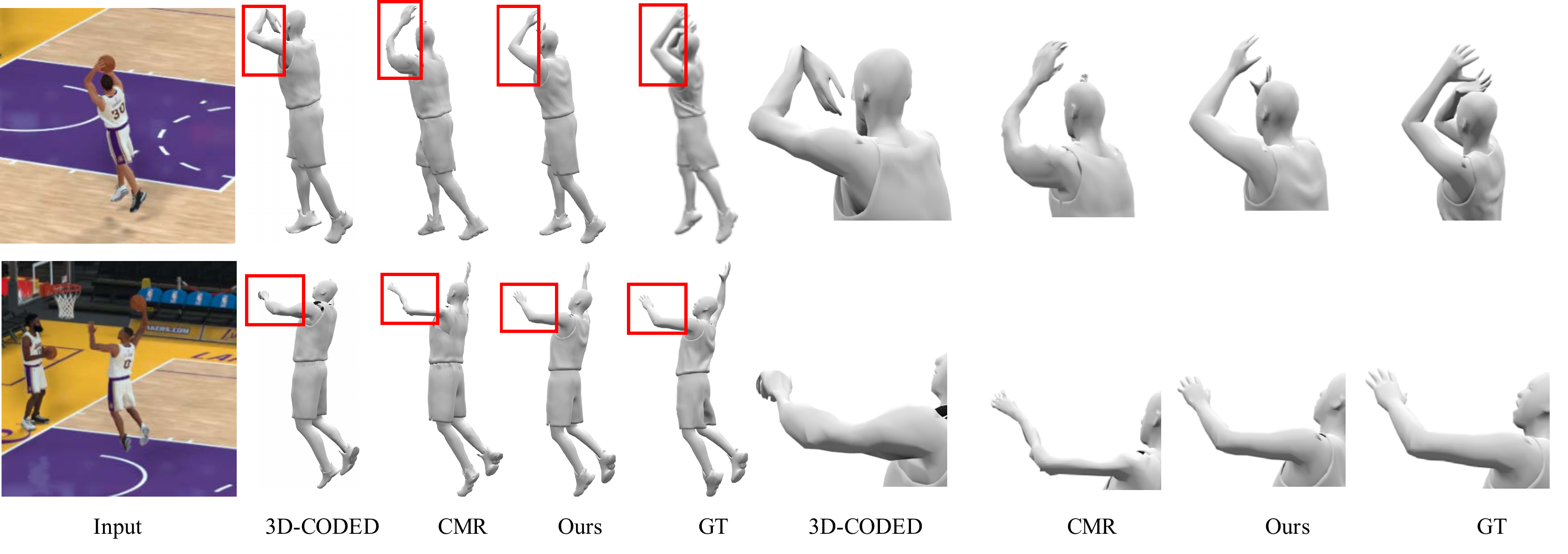}
\end{center}
  \caption{\textbf{Comparison with 3D-CODED~\cite{groueix20183d} and CMR~\cite{kolotouros2019cmr}.} 
  Column 1 is input, columns 2-5 are reconstructions in the image view, columns 6-9 are zoomed-in version of the red boxes. The baseline methods exhibit poor deformations for large deviations from the rest pose.} 
\label{geometry_learning}
\end{figure}

\noindent{\bf{Comparison with Other Geometry Learning Methods.}} Fig.~\ref{geometry_learning} compares SkinningNet with two state of the art mesh-based shape deformation networks: 3D-CODED~\cite{groueix20183d} and CMR~\cite{kolotouros2019cmr}. The baseline methods are retrained on the same data as SkinningNet for fair comparison. For 3D-CODED, we take 3D pose as input instead of a point cloud to deform the template mesh. For CMR, we only use their mesh regression network (no SMPL regression network) and replace images with 3D pose as input. Both methods use the same 3D pose encoder as SkinningNet. The input template mesh is set to the prediction of IdentityNet. Unlike baseline methods, SkinningNet does not suffer from substantial deformation errors when the target pose is far from the rest pose. Table~\ref{geometry_table} provides further quantitative results based on mean per vertex position error (MPVPE) with and without procrustes alignment.

\section{Discussion}

We have presented a novel system for state-of-the-art, detailed 3D reconstruction of complete basketball player models from single photos. Our method includes 3D pose estimation, jump estimation, an identity network to deform a template mesh to the person in the photo (to estimate rest pose shape), and finally a skinning network that retargets the shape from rest pose to the pose in the photo. We thoroughly evaluated our method compared to prior art; both quantitative and qualitative results demonstrate substantial improvements over the state-of-the-art in pose and shape reconstruction from single images.  For fairness, we retrained competing methods to the extent possible on our new data. Our data, models, and code will be released to the research community. 

\noindent{\bf{Limitations and future work}} This paper focuses solely on high quality shape estimation of basketball players, and does not estimate texture -- a topic for future work. Additionally IdentityNet can not model hair and facial identity due to lack of details in low resolution input images. Finally, the current system operates on single image input only; a future direction is to generalize to video with temporal dynamics.

\noindent{\bf{Acknowledgments}} This work was supported by NSF/Intel Visual and Experimental Computing Award \#1538618 and the UW Reality Lab funding from Facebook, Google and Futurewei. We thank Visual Concepts for allowing us to capture, process, and share NBA2K19 data for research.

%
%
\bibliographystyle{splncs04}
\bibliography{nba}

\pagebreak

\begin{center}
\textbf{\large Reconstructing NBA Players\\Supplementary Material}
\end{center}

\setcounter{section}{0}
\makeatletter

\section{NB2K Dataset Capture}
In this section we provide more details of how we select captures of the NBA2K dataset.

One way to decide which frames to capture is to let the game use its AI where two teams play against each other, however we found that the variety of poses captured in this manner is rather limited. It captures mostly walking and running people, while we target more complex basketball moves.  Instead, we have people play the game and proactively capture frames where dunk, dribble, shooting, and other complex basketball moves occur. 

\section{PoseNet}
In this section we provide more details for the PoseNet architecture and setup.

The input is a single, person-centered image with dimensions $256\times256$. We extract ResNet~\cite{xiao2018simple} features from layer 4 and supply them to four separate network branches (2D pose, 3D pose, jump class, jump height). The 2D and 3D pose branches consist of 3 set of Deconvolution-BatchNorm-ReLu blocks. For the jump class, we use a fully connected layer followed by two linear residual blocks~\cite{martinez2017simple} to get the final output and we use the same network architecture for the jump height branch. We estimate both the jump class and the jump height because the jump class can serve as a threshold to reject the inaccurate jump height prediction in the global position estimation.

The 2D pose branch outputs a set of 2D $64\times64$ heatmaps, one for every keypoint, indicating where a particular keypoint is located. Similarly, the 3D pose branch outputs a set of 2D $64\times64$ location maps~\cite{mehta2017vnect}, where each location map indicates the possible 3D location for every pixel. Each location map has 3 channels that encode the $XYZ$ position of a keypoint with respect to pelvis. To generate the ground truth heatmaps, we first transform the 2D pose from its original image resolution ($256\times256$) to $64\times64$ resolution, and then generate a 2D Gaussian map centered at each joint location. For ground truth XYZ location maps, we put the 3D joint location at the position where the heatmap has non-zero value. To obtain the final output, we take the location of the maximum value in every keypoint heatmap to get the 2D pose at $64\times64$ resolution and use it to sample the 3D pose from the $XYZ$ location maps. After that, the 2d pose is transformed to original $256\times256$ resolution. The ground truth jump height is directly extracted from the game, and the jump class is set to 1 if the jump height is greater than 0.1m.

\begin{figure}
\begin{center}
	\includegraphics[width=1.0\linewidth]{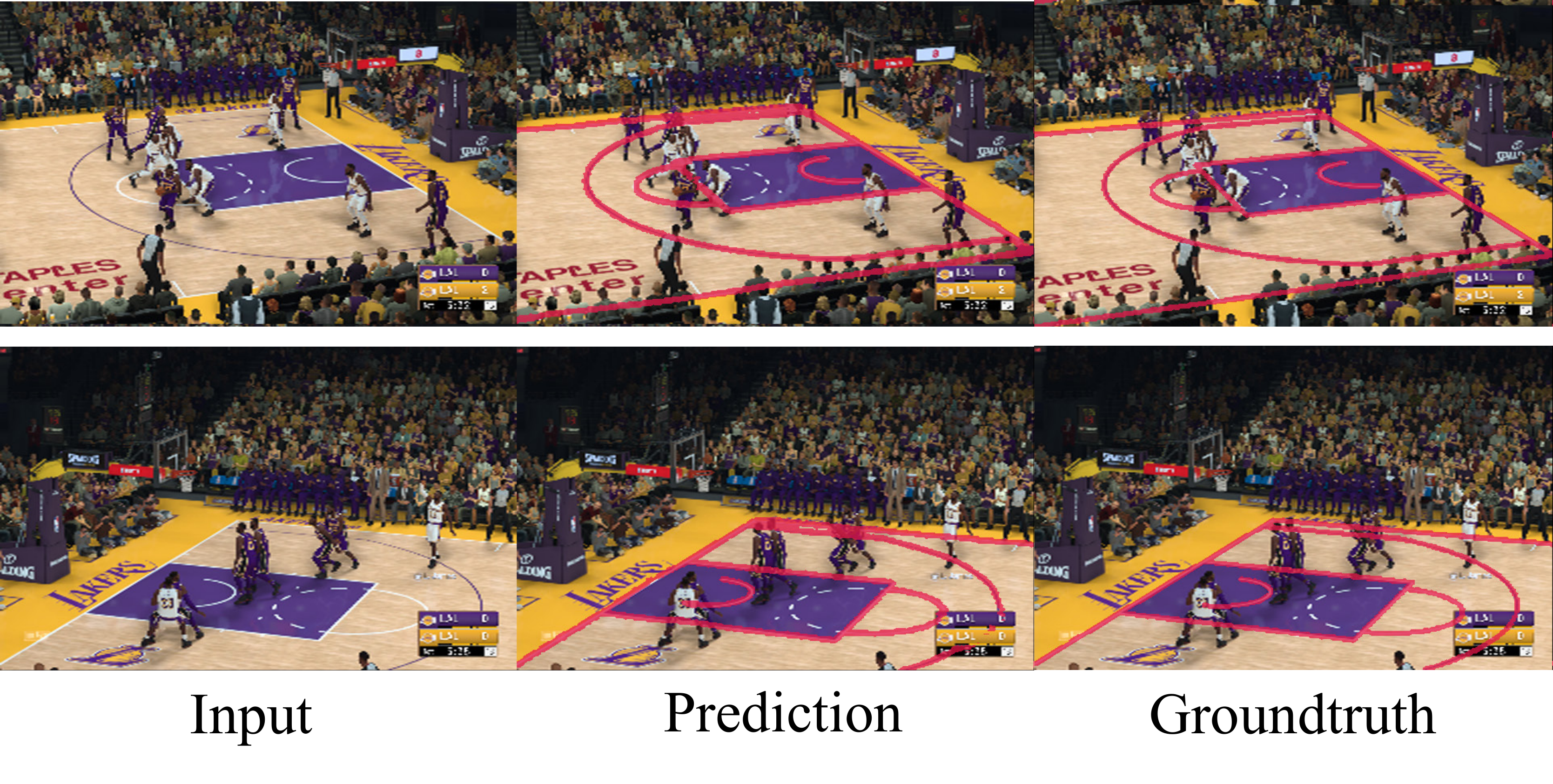}
\end{center}
  \caption{\textbf{Court line generation on synthetic data.} For every example, from left to right: input image, predicted court lines overlaid on the input image, ground truth court lines overlaid on the input image.}
\label{court_line_syn}
\end{figure}

\begin{figure}
\begin{center}
	\includegraphics[width=1.0\linewidth]{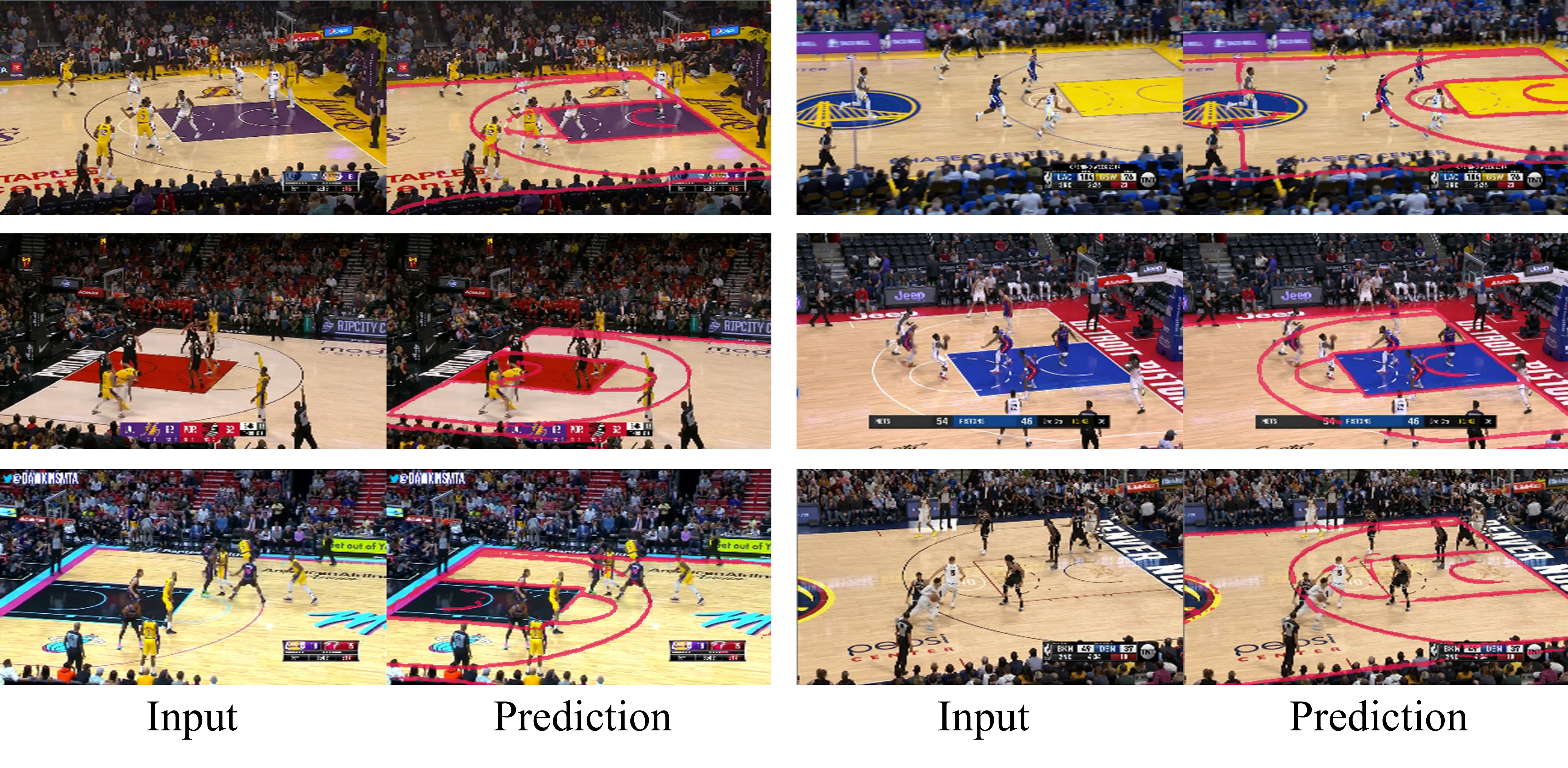}
\end{center}
  \caption{\textbf{Court line generation on real data.} For every example, left is input image, right is predicted court lines overlaid on the input image.}
\label{court_line_real}
\end{figure}

\begin{figure}
\begin{center}
  \includegraphics[width=1.0\linewidth]{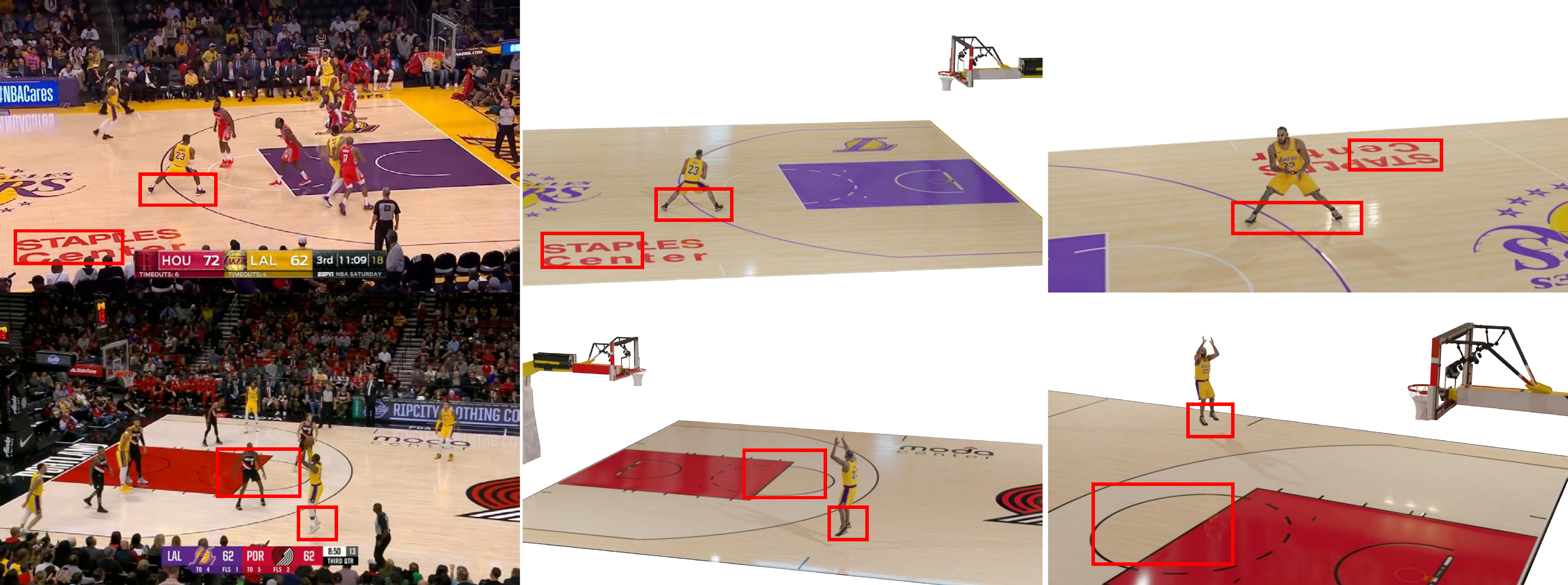}
\end{center}
  \caption{\textbf{Global position estimation. Please zoom in to see details.} From left to right: input images, two views of the estimated location (middle and right). Note the location of players with respect to court lines (marked with red boxes).}
\label{global_position}
\end{figure}

\section{Global Position}
In this section we describe the process of placing a 3D player in its corresponding position on (or above) the basketball court.

Since a basketball court with players typically has more occlusions (and curved lines) than a soccer field, we found the traditional line detection method used in~\cite{rematas2018soccer} fails. To get robust line features, we train a pix2pix~\cite{isola2017image} network to translate basketball images to court line masks. For the training data, we use synthetic data from NBA2K, where the predefined 3D court lines are projected to image space using the extracted camera parameters. To demonstrate the robustness of our line feature extraction method, we provide the results on synthetic data in Figure~\ref{court_line_syn} and real data in Figure~\ref{court_line_real}.  

After estimating the camera parameters, we place the player mesh in 3D by considering its 2D pose in the image and the jumping height (Sec 4.1):

\begin{align}
& V_{c} = \begin{bmatrix}
                        (x_{p}-p_x)\frac{z_{c}}{f} \\
                        (y_{p}-p_y)\frac{z_{c}}{f} \\
                        z_{c}
                    \end{bmatrix} \label{eq3} \\
& y_{w}=R_2\cdot(V_{c}-T) \label{eq4}
\end{align}
where $R_2$ is the second column of the extrinsic rotation matrix; T is the extrinsic translation; $f$ is focal length; $(p_x,p_y)$ is the principle point; $V_{c}$ is the camera coordinates of the lowest joint (e.g. foot); $y_{w}$ is the world coordinate $y$-component of the lowest joint, which equals the predicted jump height; $(x_{p}, y_{p})$ are the pixel coordinates of the lowest joints. Substituting Eqn. \ref{eq3} into Eqn. \ref{eq4}, we can solve for $z_c$ (camera coordinate in z-component for lowest joints), from which we can further compute the global position of the player. In Figure~\ref{global_position}, we show our results of global position estimation. We can see that our method can accurately place players (both airborne and on the ground) on the court due to accurate jump estimation.

\section{Mesh Generation}
\subsection{SkinningNet}
In this section we provide more details for the SkinningNet architecture.

As we noted in the main paper, the pose encoder is comprised of linear residual block~\cite{martinez2017simple} followed by a fully connected layer. The linear residual block consists of four FC-BatchNorm-ReLu-Dropout blocks with skip connection from the input to the output. For the mesh part, we denote Spiral Convolution~\cite{bouritsas2019neural} as SC, mesh downsampling and upsampling operator~\cite{COMA:ECCV18} as DS and US. The mesh encoder consists of four SC-ELU~\cite{clevert2015fast}-DS blocks, followed by a FC layer. The mesh decoder consists of a FC layer, four US-SC-ELU blocks, and a SC layer for final processing. We follow COMA~\cite{COMA:ECCV18} to perform the mesh sampling operation where vertices are removed by minimizing quadric errors~\cite{garland1997surface} during down-sampling and added using barycentric interpolation during up-sampling. In table~\ref{net_arch}, we provide detailed settings for the mesh encoders and decoders of different body parts.

\textit{Training details.}  For training IdentityNet and SkinningNet, we use batch size of 16 for 200 epochs and optimize with the Adam solver~\cite{kingma2014adam} with weight decay set to $5\times 10^{-5}$. Learning rate for IdentityNet is 0.0002 while learning rate for SkinningNet is 0.001 with a decay of 0.99 after every epoch. The weights of different losses are set to $\omega_{Z}=5,\omega_{mesh}=50$.

\begin{table}
\begin{center}
\begin{tabular}{c|c|c|c|c|c|c}
\hline
& head & arm & shoes & shirt & pant & leg \\
\hline
NV & 348 & 842 & 937 & 2098 & 1439 & 372 \\
\hline
DS Factor & (2,2,1,1) & (2,2,2,1) & (2,2,2,1) & (4,2,2,2) & (2,2,2,2) & (2,2,1,1)\\
\hline
NZ & \multicolumn{6}{c}{32 for all body parts} \\
\hline
Filter Size & \multicolumn{6}{c}{ (16,32,64,64) for encoders, (64,32,16,16,3) for decoders} \\
\hline
Dilation & \multicolumn{6}{c}{ (2,2,1,1) for encoders, (1,1,2,2,2) for decoders} \\
\hline
Step Size & \multicolumn{6}{c}{ (2,2,1,1) for encoders, (1,1,2,2,2) for decoders} \\
\hline

\end{tabular}
\end{center}
\caption{\textbf{Network architecture for mesh encoders and decoders of different body parts.} NV represents vertices numbers, DS factor represents downsampling factors. NZ represents the hidden size of latent vector. Filter Size represents the output channel of SC. Dilation represents dilation ratio for SC. Step size represents hops for SC.}
\label{net_arch}
\end{table}

\subsection{Combining body part meshes}
In this section, we provide details of the interpenetration optimization.

\begin{figure}
\begin{center}
	\includegraphics[width=0.6\linewidth]{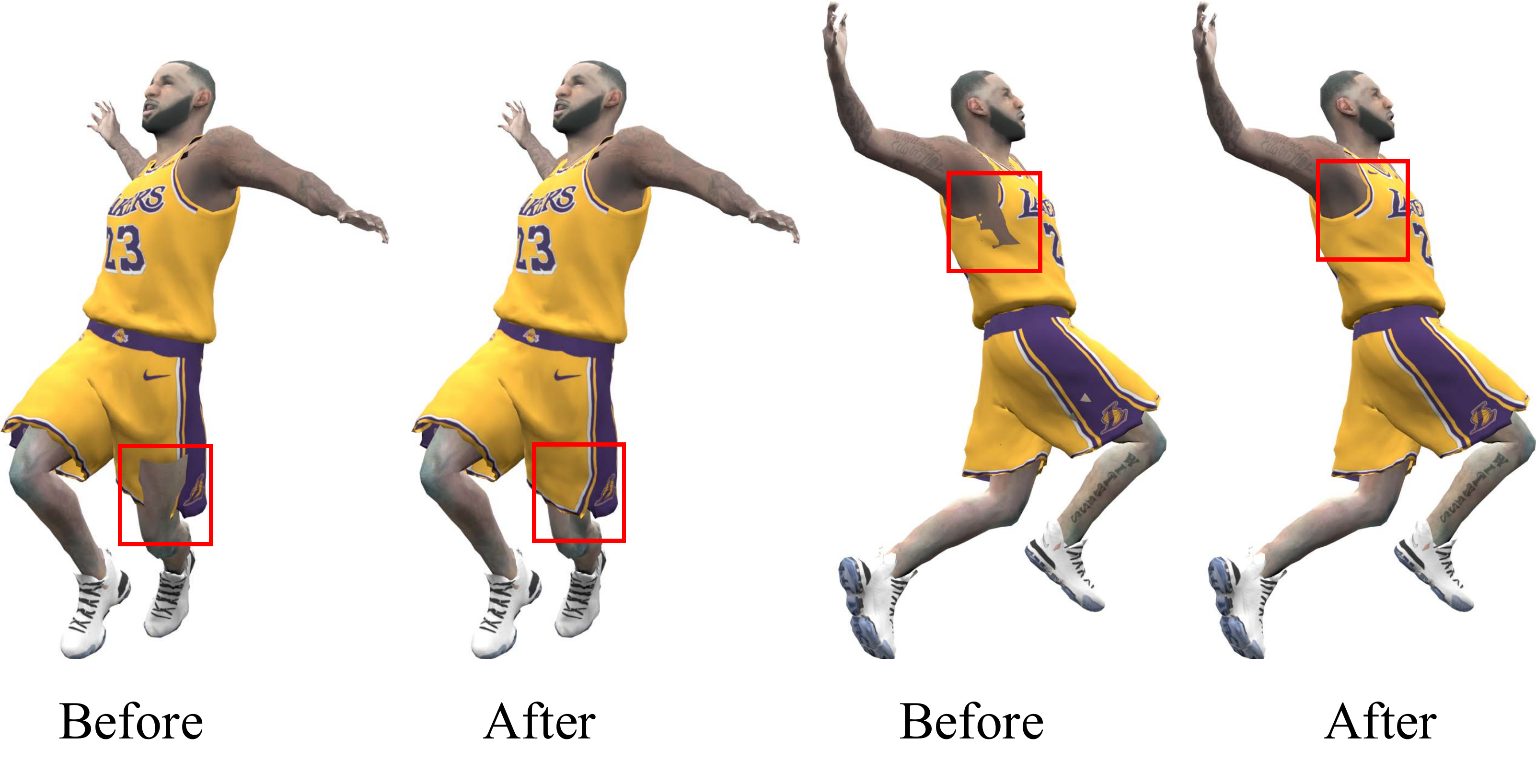}
\end{center}
  \caption{\textbf{Before and after interpenetration optimization.} Note the garment in the red square. Ground truth textures are used to better visualize the intersection.}
\label{penetration}
\end{figure}

As we noted in the main paper, we first detect all the body part vertices in collision with clothing as in~\cite{pavlakos2019smlx}, and then  follow \cite{sorkine2007rigid,sorkine2004laplacian} to deform the mesh by moving collision vertices inside the garment while preserving local rigidity of the mesh. This detection-deformation process is repeated until there is no collision or the number of iterations is above a threshold (10 in our experiments). Before each mesh deformation step, collision vertices are first moved in the direction opposite their vertex normals by 10mm. Then we optimize the remaining vertex positions of body parts by minimizing the following loss:
\begin{equation}
\mathcal{L}_{pen} = \omega_{data}\mathcal{L}_{data} + \omega_{lap}\mathcal{L}_{lap} + \omega_{el}\mathcal{L}_{el}
\end{equation}
$\mathcal{L}_{data}=\lVert {V - V^{*}} \rVert_{2}$ forces optimized vertices $V$ to stay close to the SkinningNet inferred vertices $V^{*} = V(Z_{pred})$, $\mathcal{L}_{lap} = \lVert {\Delta_{V} - \Delta_{V^*}} \rVert_{F}$ is the Frobenius norm of Laplacian difference between the optimized and inferred meshes, and $\mathcal{L}_{el} = \lVert {\frac{E}{E^*} - 1} \rVert$ encourages the optimized edge length $E$ to be same as the inferred edge length $E^{*}$. Each of these losses is taken as a sum over all vertices or edges. We set $\omega_{data}=1,\omega_{lap}=0.1,\omega_{el}=0.1$ respectively. We use an L-BFGS solver~\cite{liu1989limited}, running for 20 iterations.  Note that detected collision vertices, after being moved inward, are fixed during the optimization process. This hard constraint ensures the optimization will not move these vertices outside garments in future iterations. Figure~\ref{penetration} shows results before and after interpenetration optimization for two examples.

\begin{figure}
\vspace{-15pt}
\begin{center}
  \includegraphics[width=1\linewidth]{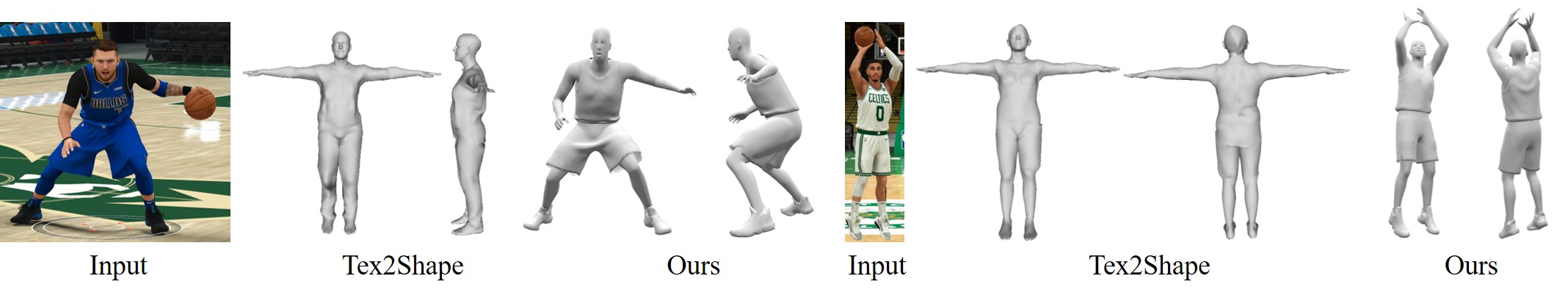}
\end{center}
\vspace{-25pt}

  \caption{\textbf{Comparison with Tex2shape\cite{alldieck2019tex2shape}}. Note that tex2shape only predicts rough body shape compared to our reconstructions. We follow their advice to select images where person is large and fully visible.}
\label{tex2shape}
\vspace{-15pt}
\end{figure}

\section{Further Qualitative Evaluation}
In this section, we provide additional qualitative comparisons that further demonstrate the effectiveness of our system.

Fig \ref{tex2shape} shows qualitative comparison with tex2shape~\cite{alldieck2019tex2shape}. Note that tex2shape is only trained with their A-pose data and directly tested on NBA images. We can see our method can generate better shirt wrinkles and body details under different poses.

\begin{figure}
\begin{center}
  \includegraphics[width=1.0\linewidth]{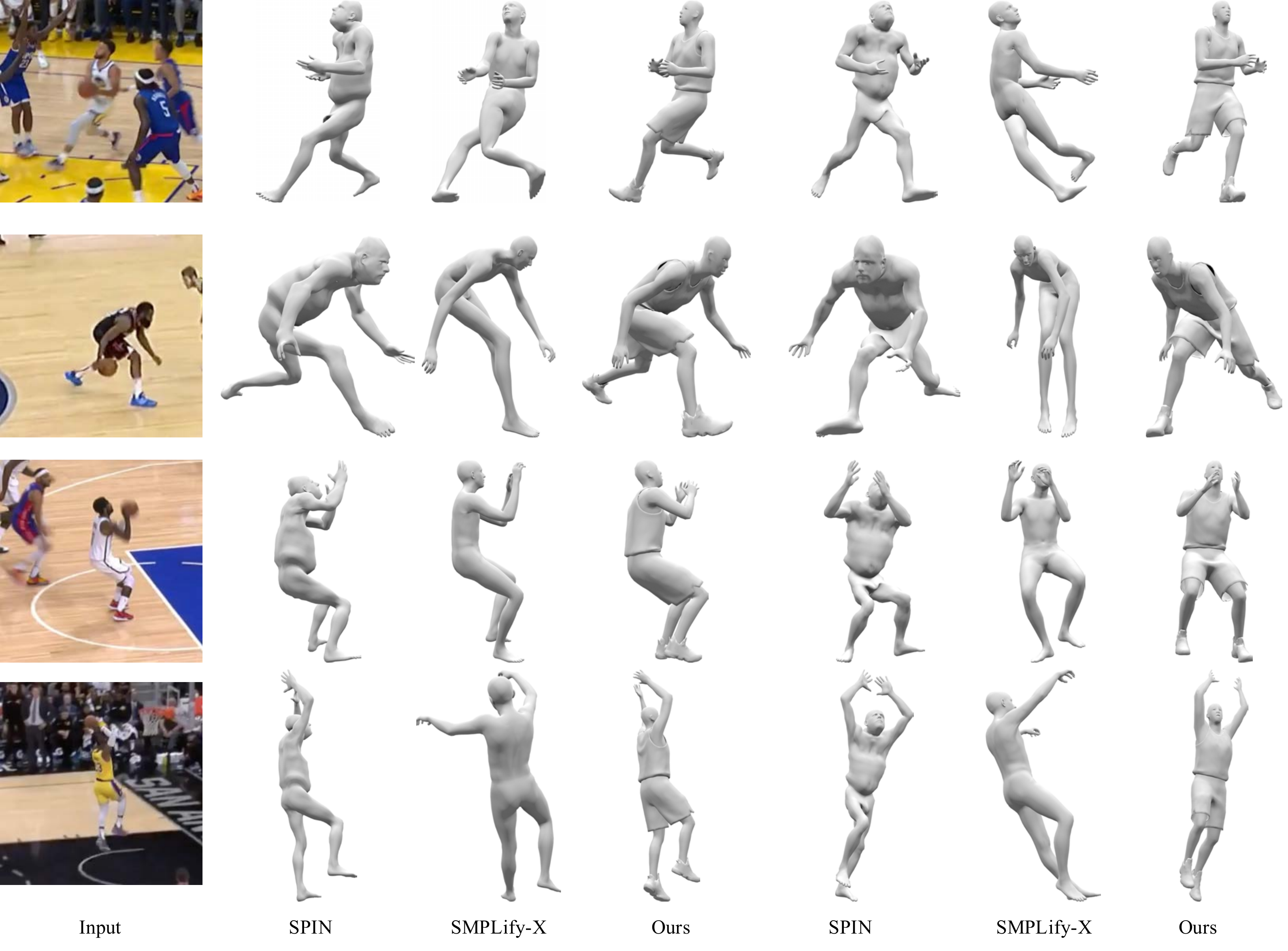}
\end{center}
  \caption{\textbf{Comparison with SMPL-based methods on real images.} Column 1 is input, columns 2-4 are reconstructions in the image view, columns 5-7 are visualizations from a novel viewpoint.  Note the significant difference in body pose between ours and SMPL-based methods; our results are qualitatively much more similar to what is seen in the input images. In addition, SMPL-based methods do not handle clothing.}
\label{smpl_real}
\end{figure}

\begin{figure}
\begin{center}
  \includegraphics[width=1.0\linewidth]{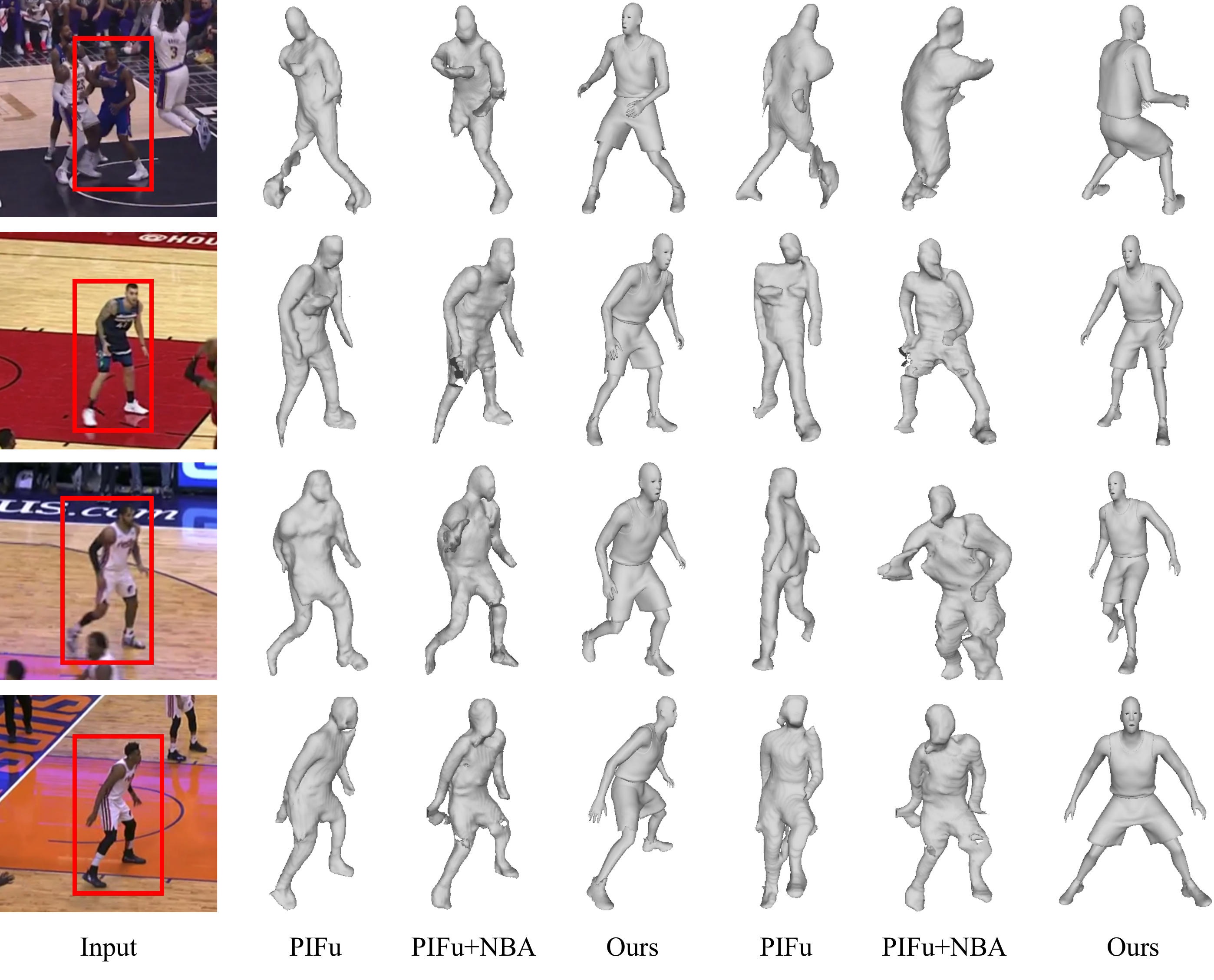}
\end{center}
  \caption{\textbf{Comparison with PIFu~\cite{pifuSHNMKL19} on real images.} Column 1 is input (red box shows the target player), columns 2-4 are reconstructions in the image view, columns 5-7 are reconstructions in a novel view. PIFu fails to reconstruct high quality human shapes from real images, even when the players are in nearly standing poses.}
\label{pifu_real}
\end{figure}

\begin{figure}
\begin{center}
  \includegraphics[width=1.0\linewidth]{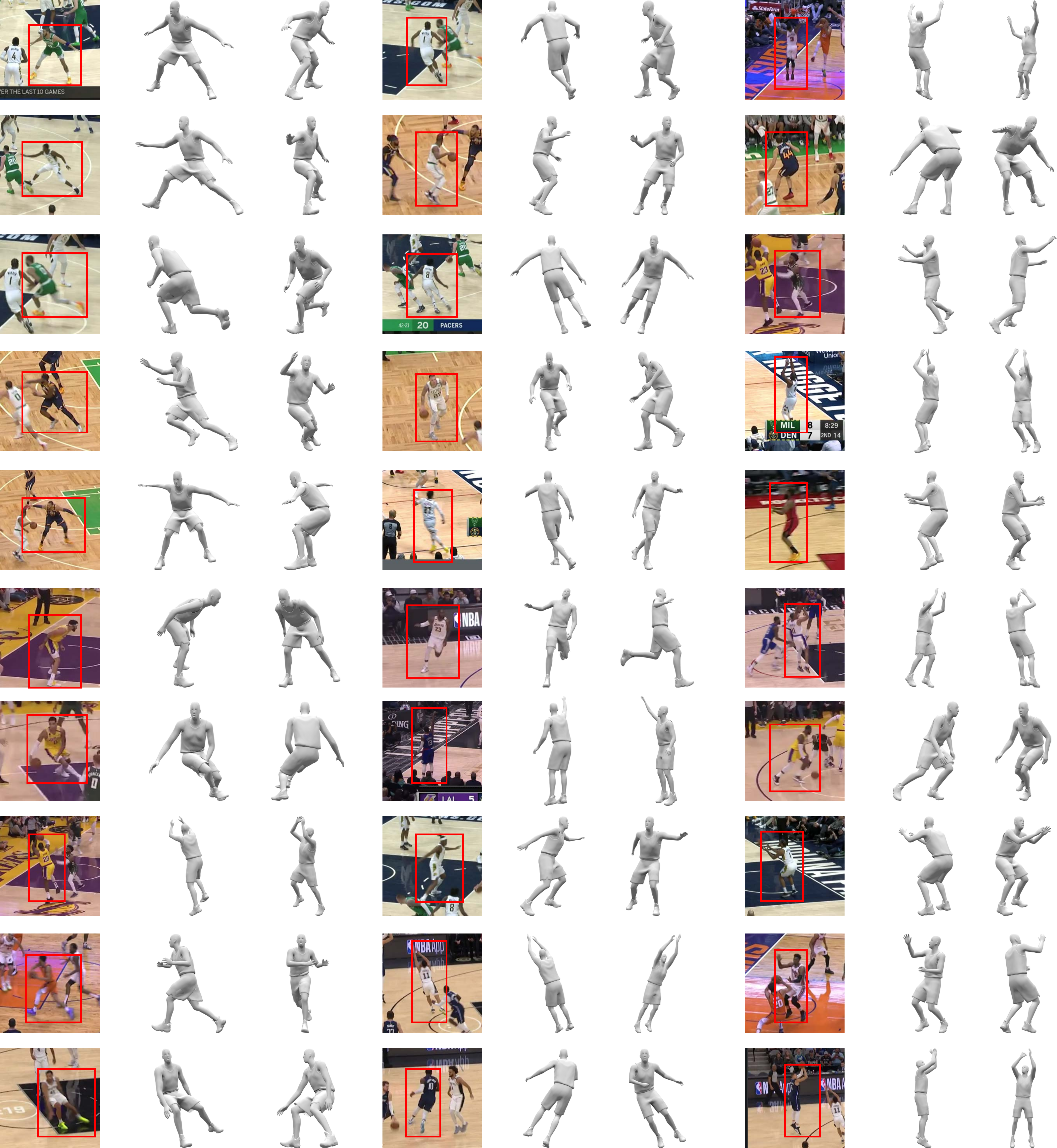}
\end{center}
  \caption{\textbf{Qualitative Results on real images. Please zoom in to see details.} For every example, left is input (red box shows the target player), middle is reconstruction in the image view, right is reconstruction in a novel view. Our method generalizes well on real images under a variety of poses.}
\label{more_real}
\end{figure}

In the main paper, we only provide qualitative comparisons for synthetic data with state-of-the-art methods. In Figure~\ref{smpl_real}, we compare our method against the best-performing SMPL-based methods~\cite{pavlakos2019smlx,kolotouros2019spin} on real images. In Figure~\ref{pifu_real}, we additionally compare with PIFu~\cite{pifuSHNMKL19}, the state-of-the-art method for clothed subjects, on real images. Our system generates more stable poses and more realistic, fine details for real images. 

In Figure~\ref{more_real}, we provide additional qualitative results of our method for real images. Our method can reconstruct 3D shape of different people under various poses on real images.

\begin{figure}
\begin{center}
  \includegraphics[width=1.0\linewidth]{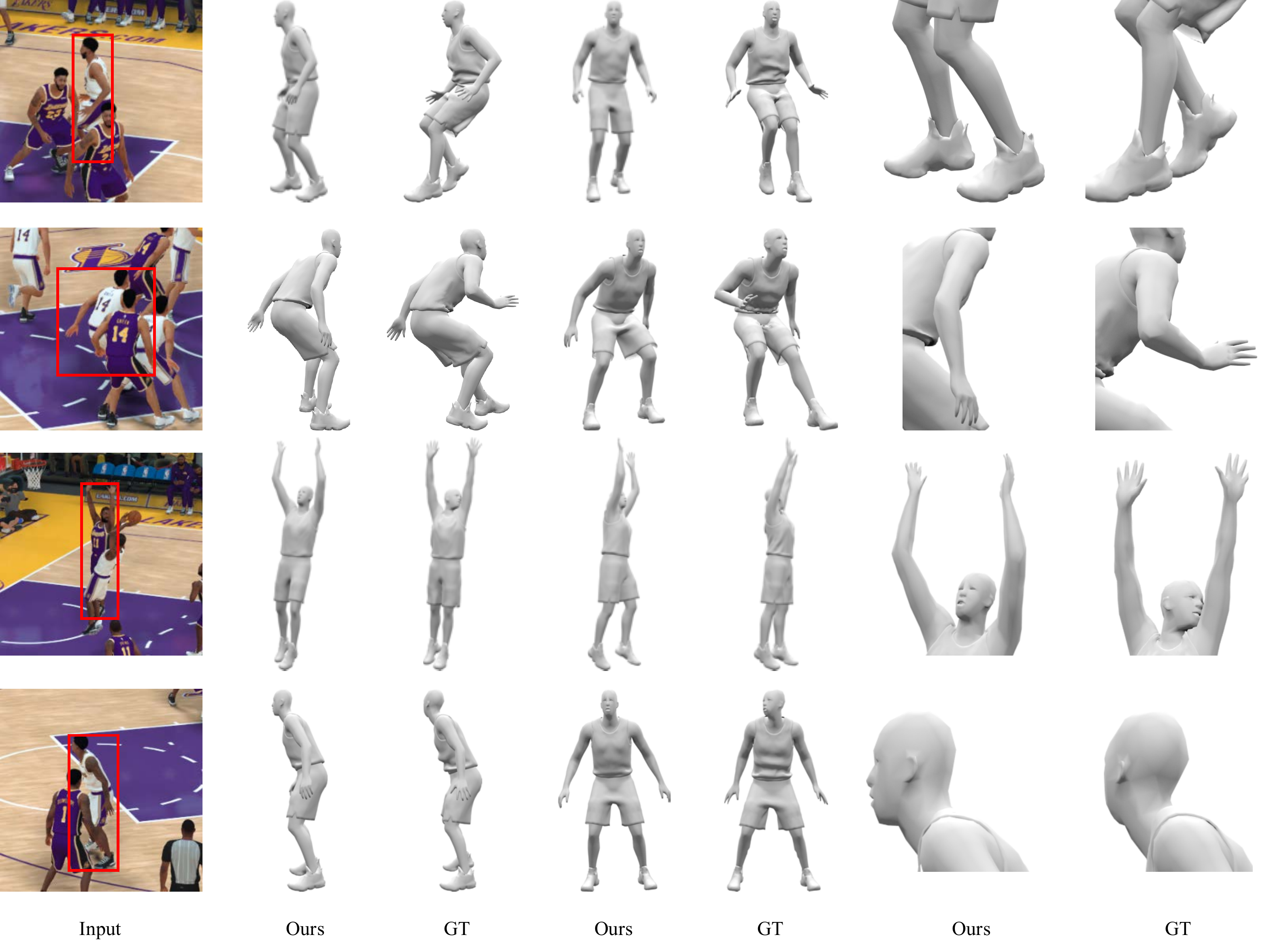}
\end{center}
  \caption{\textbf{Typical failure cases of our approach.} Column 1 is input (red box shows the target player), columns 2-3 are reconstructions in the image view, columns 4-5 are reconstructions in a novel view, columns 6-7 are zoomed-in versions of main errors. Failures include erroneous pose due to heavy occlusion in multi-person scenes (first and second example), incorrect orientation of head and hands (third and fourth example).}
\label{failure_cases}
\end{figure}

In Figure \ref{failure_cases}, we provide examples where our approach fails to reconstruct a correct 3D shape from single view images.

\end{document}